\documentclass[3p,times,procedia]{elsarticle}
\usepackage{ecrc}
\usepackage[bookmarks=false]{hyperref}
    \hypersetup{colorlinks,
      linkcolor=blue,
      citecolor=blue,
      urlcolor=blue}

\volume{00}

\firstpage{1}

\journalname{Procedia Computer Science}

\runauth{Lingxiao Meng et al.}

\jid{procs}

\usepackage{amssymb}

\usepackage{multirow}
\usepackage{subfigure}
\usepackage{cleveref}
\usepackage[figuresright]{rotating}

\begin{document}
\begin{frontmatter}

\dochead{Proceedings of International Conference on Biomimetic Intelligence and Robotics}

\title{Virtual Reality Based Robot Teleoperation via Human-Scene Interaction}

\author[a,b]{Lingxiao Meng\eqref{c1}}
\author[a,b]{Jiangshan Liu\eqref{c1}}
\author[d,e]{Wei Chai}
\author[a,b,c]{Jiankun Wang\corref{cor1}}
\author[a,b]{Max Q.-H. Meng\corref{cor1}}

\address[a]{Shenzhen Key Laboratory of Robotics Perception and Intelligence, Shenzhen, China}
\address[b]{Department of Electronic and Electrical Engineering, Southern University of Science and Technology, Shenzhen, China}
\address[c]{Jiaxing Research Institute, Southern University of Science and Technology, Jiaxing, China}
\address[d]{Senior Department of Orthopedics, the Fourth Medical Center of PLA General Hospital, Beijing, China}
\address[e]{National Clinical Research Center for Orthopedics, Sports Medicine and Rehabilitation, Beijing, China}

\begin{abstract}
Robot teleoperation gains great success in various situations, including chemical pollution rescue, disaster relief, and long-distance manipulation. In this article, we propose a virtual reality (VR) based robot teleoperation system to achieve more efficient and natural interaction with humans in different scenes. A user-friendly VR interface is designed to help users interact with a desktop scene using their hands efficiently and intuitively. To improve user experience and reduce workload, we simulate the process in the physics engine to help build a preview of the scene after manipulation in the virtual scene before execution. We conduct experiments with different users and compare our system with a direct control method across several teleoperation tasks. The user study demonstrates that the proposed system enables users to perform operations more instinctively with a lighter mental workload. Users can perform pick-and-place and object-stacking tasks in a considerably short time, even for beginners. Our code is available at \url{https://github.com/lingxiaomeng/VR_Teleoperation_Gen3}.
\end{abstract}

\cortext[cor1]{Corresponding author.}
\eqtext[c1]{indicates equal contributions}

\end{frontmatter}

\email{wangjk@sustech.edu.cn; max.meng@ieee.org}

\section{Introduction}

Robot teleoperation, referring to operating robots remotely, has proven valuable in various scenes, including chemical pollution rescue, disaster relief operations, and long-distance manipulation \cite{i1,i2}. When physical access to the robot's workspace is hard or even impossible, teleoperation enables professionals to execute their duties remotely. Recent advancements in robotics have resulted in the development of robots with increased degrees of freedom, facilitating the performance of intricate tasks. However, conventional interfaces such as keyboards, joysticks, and screens do not effectively aid users in controlling robots, affecting the widespread application of teleoperation.

Virtual Reality (VR) has become a progressively intuitive method for robot control, facilitating direct immersion in a virtual three-dimensional environment \cite{i3}. By replicating the robots' workspace in this way, users can have a more natural comprehension of the workspace, resulting in informed decision-making. Therefore, VR has significant potential as an interface for teleoperation, but accomplishing specific tasks via teleoperation using a VR interface is still an open problem.

Recent research in robot teleoperation has predominantly focused on robot-centric control strategies. Early researchers primarily devoted their efforts to specifying the continuous trajectory of the end effector \cite{r1}. This method requires a low-latency connection between the users and the robot, imposing significant requirements on hardware capabilities and network connectivity. In recent years, to mitigate these requirements, researchers have shifted towards enabling users to control only the end effector \cite{r6}, including controlling the end effector directly \cite{r5,r8,i4} or specifying discrete waypoints \cite{r4} to perform operations. However, these methods still burden the user to manually determine each action of the robot, which is a challenging and cumbersome task. Regardless of the inherent difficulty of the task itself, these methods heavily focus on the robot's movements. Consequently, users not only face a considerable workload during actual operation but also need a comprehensive understanding of robot kinematics and dynamics, which affects the widespread application of teleoperation.

In contrast, our system aims to develop a task-centric VR teleoperation system that emphasizes the task at hand, disregarding the manipulation of robotic hardware. Similar task-driven teleoperation has been explored in the context of traditional control interfaces \cite{r7, r9}. However, these methods still need to improve on the unintuitive nature of the operation. The same method has also been studied in VR, but the task is limited to grasping \cite{i5}. Our work addresses the challenges of providing intuitive and user-friendly teleoperation while minimizing mental workload. We contribute to developing a VR interface that enhances the users' experience through an intuitive workspace and user-friendly interactions. By perceiving the reconstructed and visualized environment within the VR device, the user can directly manipulate, grasp, and position familiar objects using their hands. Moreover, we have incorporated a physics engine that simulates the impact of the operator's actions on the environment, allowing users to preview the resulting changes in the scene.

Our contributions primarily encompass two aspects: 
1) the comprehensive implementation of the entire system; 
2) integrating a physics engine into the VR interface, enabling users to preview the operations and their effects. 

The remaining sections of this article are organized as follows: In Section 2, the composition of the whole system and the functions of each part is introduced. Section 3 introduces the methods and indicators to measure the system's performance, and Section 4 presents the performance results of the above metrics in experiments and data analysis. At last, we give a conclusion in Section 5.

\section{System Overview}

   \begin{figure}[htbp]
      \centering
      \includegraphics[width=\columnwidth]{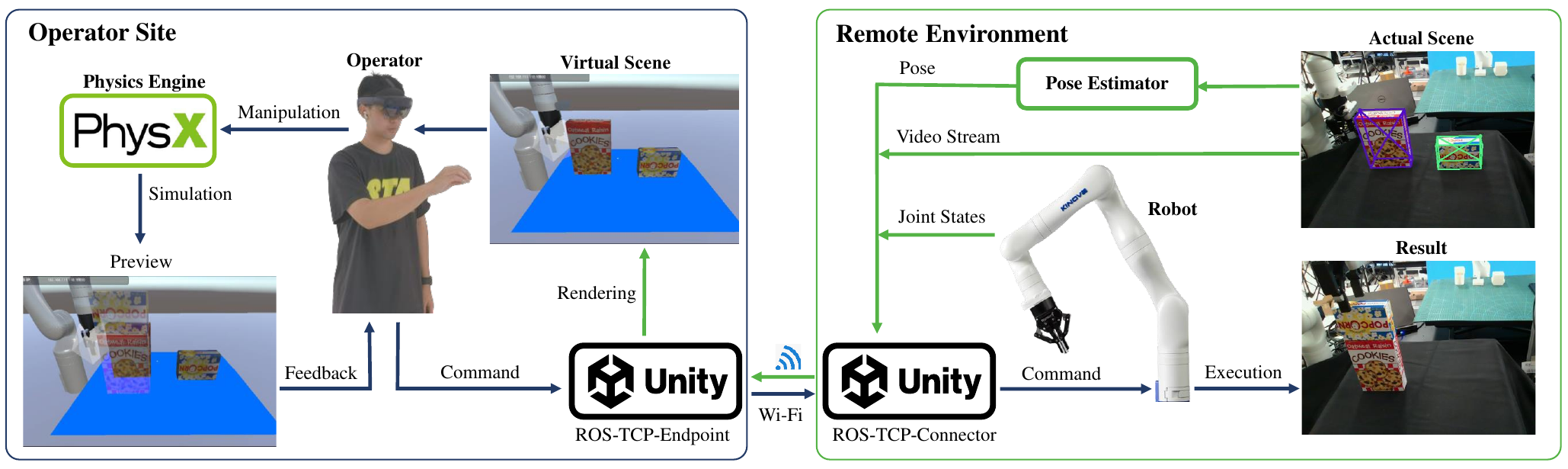}
      \caption{System structure and workflow.}
        \label{f1}
   \end{figure}

Fig. \ref{f1} illustrates the system structure and workflow. First, the VR interface provides the user with a virtual robotic arm and a desktop that accurately corresponds to the real world. Subsequently, the pose estimation algorithm synchronizes the object's pose in the VR interface with its real-world counterpart, allowing the users to manipulate the object within the virtual environment. Once the manipulation is completed, the physics engine simulates the movement of the virtual object to achieve a more precise target pose. Finally, based on the initial and target poses of the virtual object, the robotic arm automatically executes the operation on the real object.
 
In detail, the proposed system contains four components: \textit{Devices and Development}, \textit{Desktop Detection and Pose Estimation}, \textit{VR Interface}, and \textit{Robot Control}.

 \subsection{Devices and Development}

Our system utilizes Microsoft's HoloLens 2 mixed reality (MR) device, the Intel RealSense L515 depth camera, the 7 degrees-of-freedom (DoF) Kinova Gen3 robotic arm, a desktop computer, and a TP-Link router. This study primarily focuses on utilizing the VR functionality of the HoloLens 2. The end effector of the Kinova robotic arm is a two-fingered gripper known as the Robotiq 2f-85. The relative pose between the camera and the robotic arm is obtained through eye-to-hand calibration using MoveIt Calibration Tools. The desktop computer is equipped with an Intel i7-13900 CPU, 32GB RAM, and an Nvidia RTX 4090 graphics card. It is connected to the router using a network cable, while the HoloLens 2 is connected via Wi-Fi.

The development part is divided into HoloLens 2 development and ROS server development. For HoloLens 2 development, we utilize Unity 2021 to develop the VR program written in C\#. The program uses the Mixed Reality Toolkit to create an interactive system. The ROS server development includes functionalities for robotic arm control and image processing. HoloLens 2 connects with the ROS server's ROS-TCP-Endpoint node through the ROS-TCP-Connector for data transmission.
 
\subsection{Desktop Detection and Pose Estimation}

The scene of this study involves manipulating objects on a desk. To accurately simulate the collision interactions between the desktop and the objects, obtaining a physical model of the desktop is necessary. Initially, all the planes in the point cloud are segmented with Point Cloud Library (PCL) \cite{pcl}. Subsequently, the clustering algorithm removes the scattered points in the largest plane, and the largest plane's boundary is extracted to generate a mesh representation. The desktop detection functionality is integrated into the system as the \textit{DesktopDetection} service. When the service is called, it provides the mesh representation of the desktop, which is then displayed in the virtual scene.

To simplify the acquisition of 3D models of objects, we utilize the objects in Household Objects for Pose Estimation (HOPE) \cite{hope} dataset for manipulation. The HOPE dataset consists of 28 commonly encountered 3D scanned models of household objects, widely used in robot grasping tasks. Deep Object Pose Estimation (DOPE) \cite{dope} is employed for estimating the 6 DoF pose of known objects with respect to the camera optical frame using color images.

DOPE is integrated into the system as a ROS node that provides real-time publishing of the detected objects' IDs and their corresponding 6 DoF poses through \textit{object\_poses} topic.

\subsection{VR Interface}

  \begin{figure}[htbp]
      \centering
      \includegraphics[width=0.95\columnwidth]{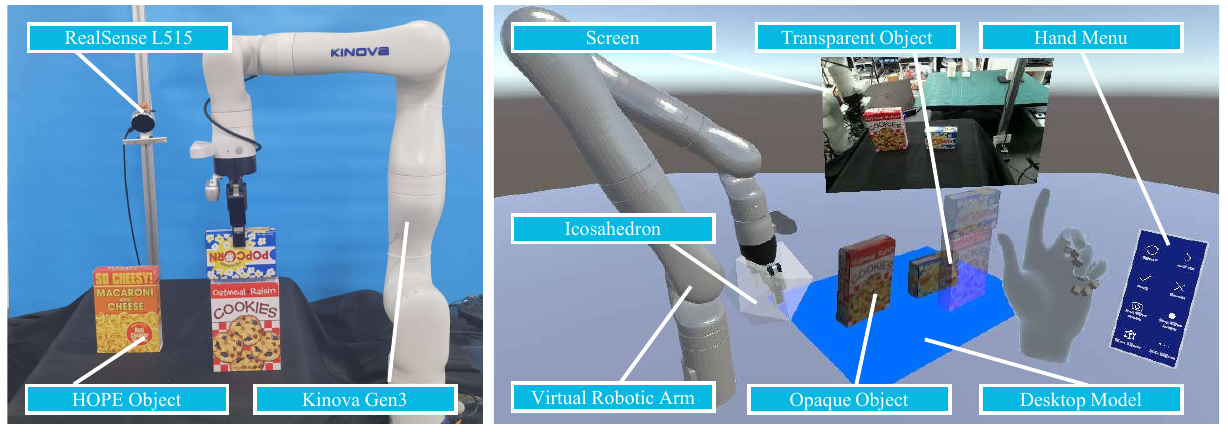}
      \caption{Experimental platform and VR interface screenshot.}
            \label{f2}

   \end{figure}

\subsubsection{Visual Interface}

As shown in Fig. \ref{f2}, the virtual scene contains the following seven visual elements. Using the virtual Kinova Gen3 robotic arm model imported by the URDF Importer, \textbf{(a) the virtual robotic arm} synchronizes with the real robotic arm by receiving \textit{joint\_states} topic. There is also \textbf{(b) the desktop model} from calling the \textit{DesktopDetection} service. In the virtual scene, the object models in the HOPE data set are imported in advance and hidden by default. After receiving the \textit{object\_poses} message, the corresponding \textbf{(c) opaque object models} are displayed according to their ids and corresponding poses. In addition, each object will have a corresponding \textbf{(d) transparent object} that displays the target state of the object that the user wants to move, which by default coincides with the opaque model. There is also a \textbf{(e) screen} to display the L515 color images received from the \textit{l515\_image\_raw} topic. An \textbf{(f) icosahedron} on the end effector of the robotic arm is attached to the tool frame of the robotic arm, through which the user can control the robotic arm. When the user extends their left hand towards themselves,  a \textbf{(g) hand menu} will be presented, allowing the user to execute specific commands by clicking the corresponding buttons.

\subsubsection{Control Interface}

  \begin{figure}[htbp]
      \centering
      \includegraphics[width=0.95\columnwidth]{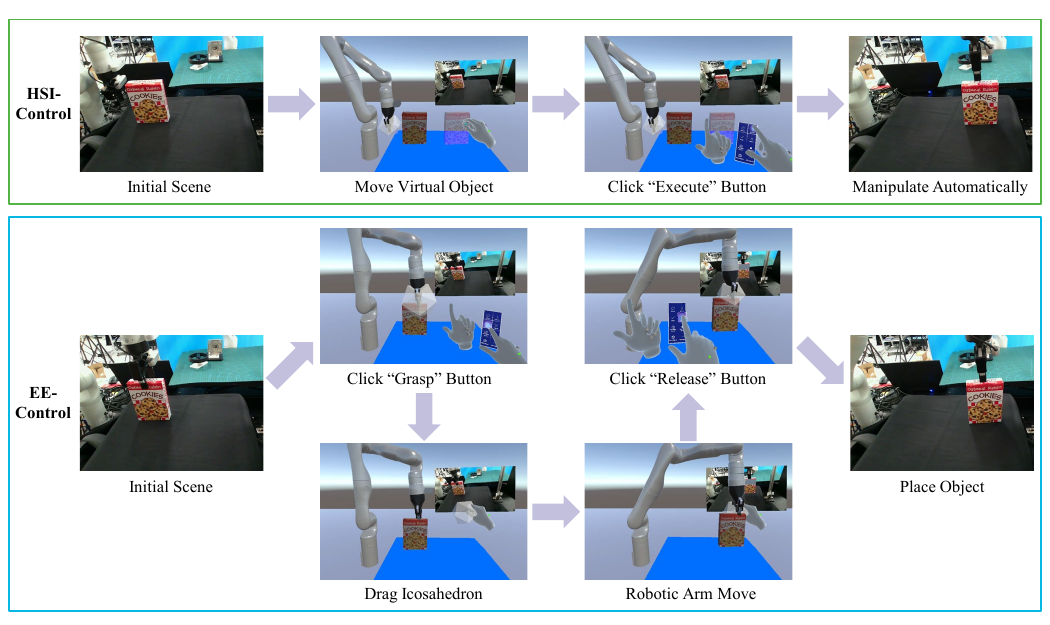}
      \caption{Operation flow of HSI-Control and EE-Control.}
            \label{f5}

   \end{figure}

As shown in Fig. \ref{f5}, the control interface consists of two modes: human-scene interaction control (HSI-Control) and traditional end effector control (EE-Control).

In the HSI-Control mode, we have designed an interface that enables users to manipulate the robotic arm by directly interacting with the scene. The user initiates the control process by grasping the virtual object that they want to move. At this point, the opaque object remains stationary while the transparent object moves in accordance with the user's hand movements. The user can then move the transparent object to the desired position and release it. Upon release, the script activates the gravity property of the transparent object. With the Unity physics engine simulation, the transparent object undergoes free fall and collides with the desktop. Finally, the transparent object smoothly lands on the desktop, assuming its final target pose. By incorporating physical simulation, the system enhances the accuracy in predicting the manipulated object's final pose, thereby improving the success rate of teleoperation. Once the user completes object manipulation, they can click the ``Execute" button in the Hand menu to command the robotic arm to perform the corresponding task. This action calls the \textit{ExecuteTask} service, which receives a request message containing the ID, initial pose, and target pose of each object in the order of the user's operations.

In the EE-Control mode, users can directly manipulate the robotic arm by dragging the icosahedron. Upon initially holding the icosahedron, its position becomes unbound from the tool frame, allowing users to move or rotate the icosahedron freely. The resulting pose of the icosahedron is then published to the control node through the \textit{target\_pose} topic, enabling real-time movement of the robotic arm synchronized with the icosahedron. Additionally, users can control the opening and closing of the gripper by clicking the ``Grasp" and ``Release" buttons in the hand menu. Each button click calls the respective \textit{GraspService} and \textit{ReleaseService}.

\subsection{Robot Control}

We develop the control node based on the official driver for the Kinova Gen3 robotic arm. When the \textit{ExecuteTask} Service is called from the HSI-Control mode, the robotic arm sequentially operates each object to complete the desired manipulation. For each object, the robotic arm moves above it, performs a top-down grab, moves it to the target position, and then releases the gripper to place the object.

In the EE-Control mode, the node puts the pose in a fixed-length queue upon receiving the \textit{target\_pose} message. The main program cyclically attempts to pop the pose from the queue, and when it gets the pose, controls the robotic arm to move to the corresponding pose.

Since the Robotiq 2f-85 gripper does not have a force sensor, two control modes are designed to operate the gripper. The first is position control, where the \textit{ReleaseService} is called to open the gripper fully. The second mode is grasp control based on motor current feedback. Upon calling the \textit{GraspService}, the gripper closes at a speed of 0.1 m/s while monitoring the motor current. If the current exceeds 0.12 A, it indicates successful object grasping, then the motor speed is set to zero.

\section{Evaluation}

\begin{figure}[ht]
  \centering
  \includegraphics[width=0.75\columnwidth]{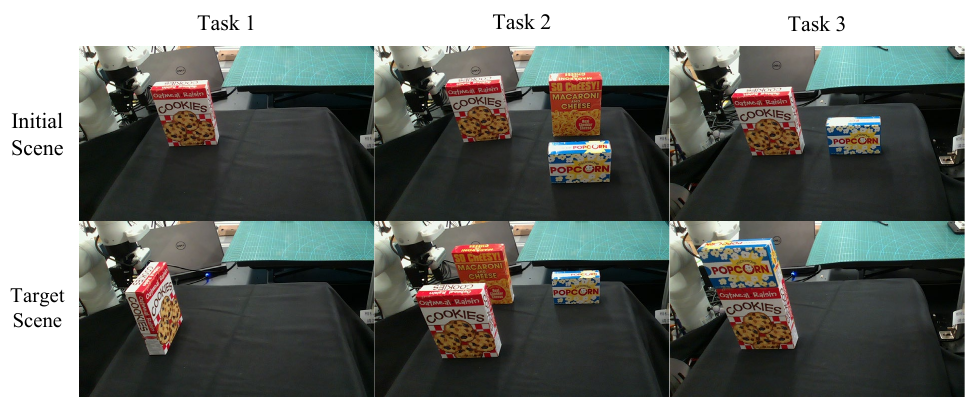}
  \caption{Participants are required to use HSI-Control and EE-Control to complete all tasks, that is, moving objects in the scene from the initial pose to the target pose.}
    \label{f3}

\end{figure}

To evaluate the effectiveness of the proposed system, we conduct real-world experiments to compare the HSI-Control with the EE-Control in three tasks, as shown in Fig. \ref{f3}. Task 1 involves a simple pick-and-place operation, Task 2 focuses on rearranging multiple objects, and Task 3 requires stacking one object on top of another.
\subsection{Participants}

The experiment involves recruiting 5 participants from the SUSTech RPAI Lab, consisting of 4 males and 1 female, aged between 21 and 28 years. All participants possess knowledge of robotics, with three of them having no prior experience in VR or MR.

\subsection{Experimental Procedure}

The experimental procedure begins by instructing the participants to wear the HoloLens 2. Subsequently, the fundamental functions and gesture operation methods of the HoloLens 2 are introduced to them. Prior to each task, the task scene is prepared in advance, and the participants are presented with an image illustrating the target scene. The participants are informed that their goal is to complete each task with speed and precision while avoiding collisions. Subsequently, tasks 1, 2, and 3 are performed sequentially, employing two distinct control methods. To mitigate potential learning bias, we randomize the order of the two methods. Upon completing all tasks, the participants are invited to fill out a written questionnaire to provide feedback on their experience.

\subsection{Measurements}

\subsubsection{Objective Indicators}

The objective measures for each task consist of \textit{completion time}, \textit{interaction time}, and \textit{success rate}. In the HSI-Control, where the robotic arm autonomously performs part of the task, the \textit{interaction time} is shorter than the \textit{completion time}. Specifically, for the HSI-Control, the \textit{completion time} of the task spans from the participant's initial interaction in VR to the moment the robotic arm completes all operations, while the \textit{interaction time} refers to the duration of the participant's VR interaction. In the EE-Control, these two terms align with the interval between the participant's initial control of the robotic arm and the task's completion. If a significant collision occurs during task execution, impeding the ability to complete the task through teleoperation, it is classified as a failure.

\subsubsection{Subjective Indicators}

The workload of the participants is measured using NASA-Task Load Index (NASA-TLX) \cite{nasa}. NASA-TLX evaluates the workload across six dimensions: Mental Demand, Physical Demand, Temporal Demand, Performance, Effort, and Frustration. Participants are required to assign scores to each dimension, ranging from 0 to 100, where lower values indicate a lower workload.

\section{Results}

\begin{figure}[ht]
    \centering
    \subfigure[]{\label{f41}\includegraphics[width=.45\columnwidth]{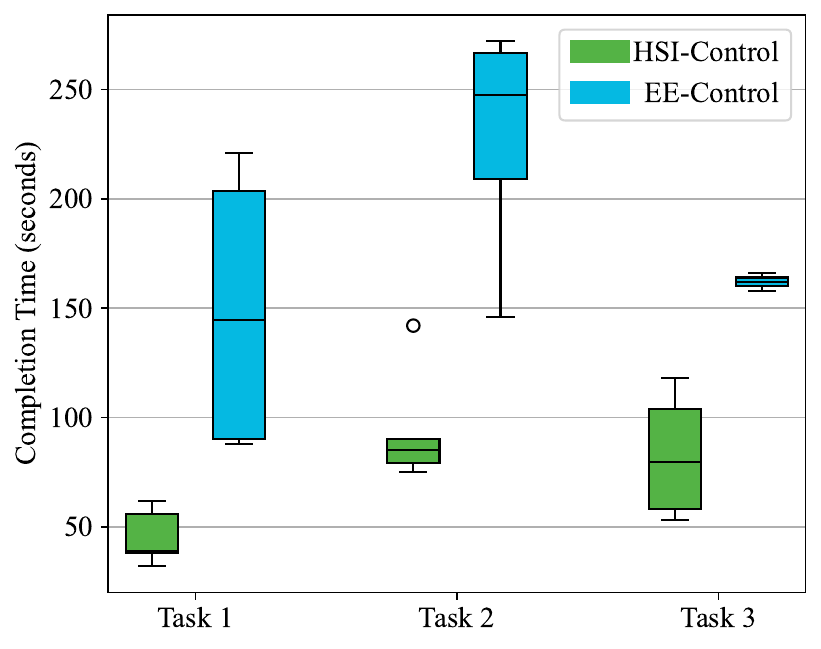}}
    \subfigure[]{\label{f42}\includegraphics[width=.45\columnwidth]{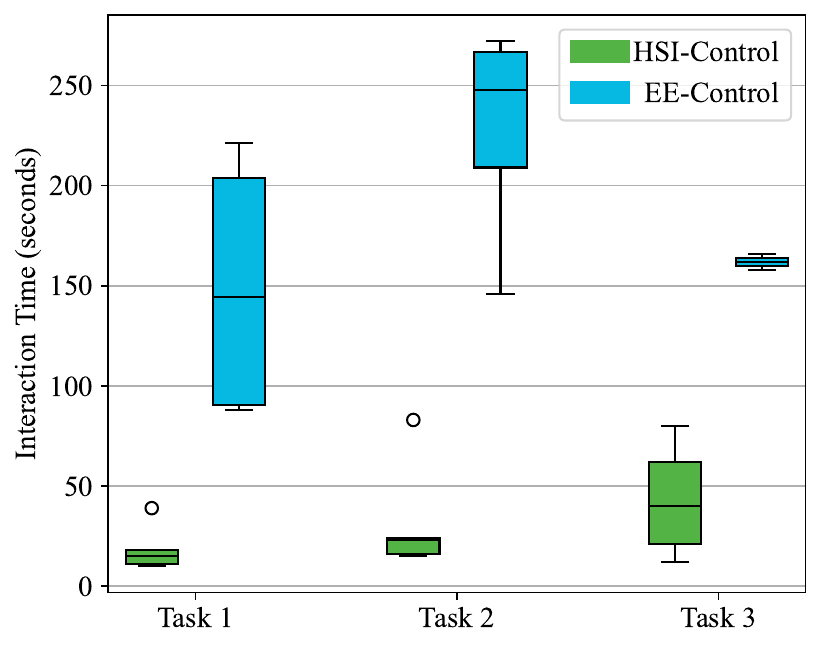}}
    \subfigure[]{\label{f43}\includegraphics[width=.45\columnwidth]{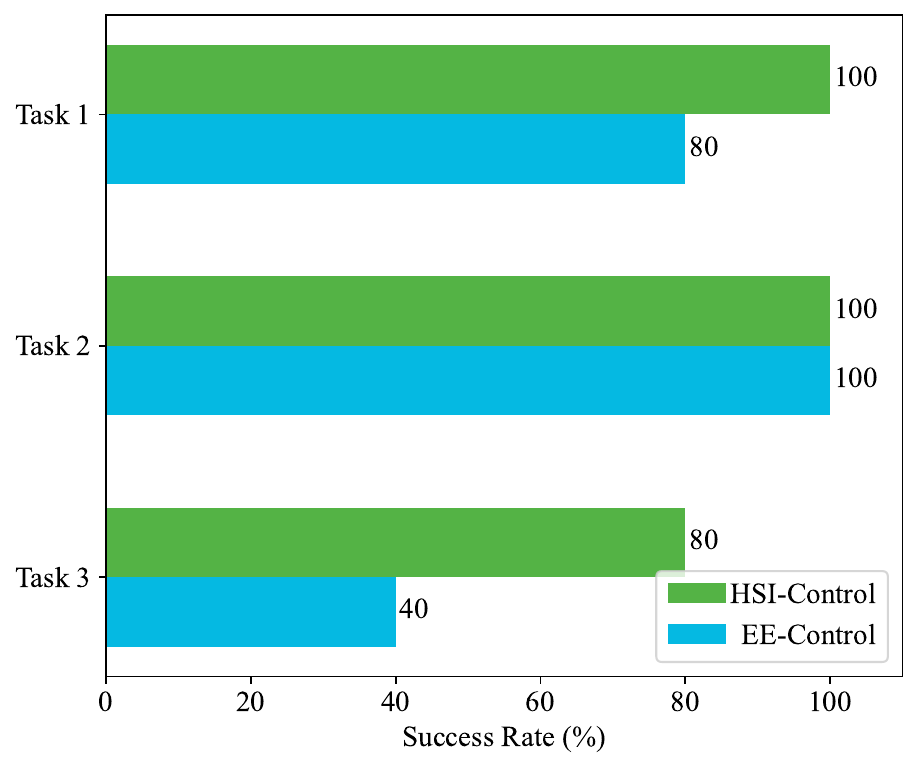}}
    \subfigure[]{\label{f44}\includegraphics[width=.45\columnwidth]{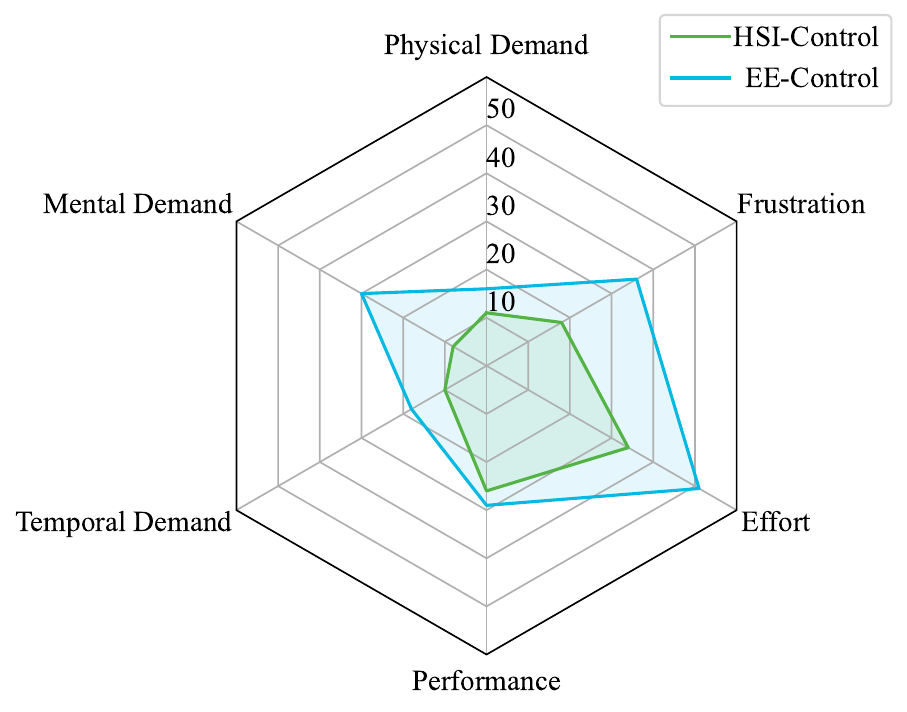}}
    \caption{(a) Comparison of \textit{completion time}; (b) Comparison of \textit{interaction time}; (c) Comparison of \textit{success rate}; (d) Comparison of workload.}
        \label{fig:4}

\end{figure}

\subsection{Objective Indicators}

Fig. \ref{f41} to \ref{f43} present the \textit{completion time}, \textit{interaction time}, and \textit{success rate} of the two methods. Across the three tasks, the HSI-Control exhibits significantly shorter \textit{completion time} ($p_{Task 1}=0.006, p_{Task 2}<0.001, p_{Task 3}=0.014$) and \textit{interaction time} ($p_{Task 1}=0.002, p_{Task 2}<0.001, p_{Task 3}=0.003$) compared with the EE-Control. Moreover, the HSI-Control demonstrates a higher \textit{success rate}, particularly in Task 3 involving object stacking. Due to the low fault tolerance during object placement in Task 3, users using the EE-Control need continuously fine-tune their actions based on the image. In contrast, the HSI-Control enables users to preview object stability in real time, replacing the fine-tuning with physical simulation. Consequently, the success rate of the task is improved.

\subsection{Subjective Indicators}

The results obtained from the NASA-TLX questionnaire are presented in Fig. \ref{f44}. The workload of HSI-Control is significantly less than that of EE-Control ($p=0.021$). The HSI-Control exhibits significantly lower levels of Mental Demand ($p=0.002$), Effort ($p=0.040$), and Frustration ($p=0.016$) compared to the EE-Control.

\subsection{Analysis}

Overall, HSI-Control demonstrates greater efficiency than EE-Control. In EE-Control, a considerable amount of time is devoted to adjusting the object's pose. In HSI-Control, this process is considerably shorter as participants typically succeed in placing the virtual object in a single attempt. Moreover, the participants report that HSI-Control exhibits greater intuitiveness and ease of operation. Furthermore, participants adapt to HSI-Control more quickly than to EE-Control in the experiment, suggesting it is more friendly to beginners.

\section{Conclusions}

We present a remote operating system that utilizes human-scene interaction for semi-known scenes. The experimental results demonstrate that the proposed method enhances efficiency and success rate and reduces user workload.

Given that our method relies on having a 3D model of the object, it is necessary to pre-model the object when dealing with unknown objects in the scene. In future work, we aim to incorporate the ``Segment Anything in NeRF" \cite{nerf1, nerf2} technique to enable HSI-Control in any scene. Additionally, our current method does not consider reorientation, and we plan to integrate a regrasp planner \cite{regrasp} into the system to incorporate this functionality.

\section*{Acknowledgements}

This work is supported by the Shenzhen Key Laboratory of Robotics Perception and Intelligence under Grant ZDSYS20200810171800001, Shenzhen Outstanding Scientific and Technological Innovation Talents Training Project under Grant RCBS20221008093305007, and National Natural Science Foundation of China grant \#62103181.

\bibliography{refs}
\bibliographystyle{elsarticle-num-names}

\end{document}